\documentclass{article}
\usepackage{spconf,amsmath,graphicx,hyperref}

\makeatletter
\let\originalthebibliography\thebibliography
\renewcommand{\thebibliography}[1]{%
  \originalthebibliography{#1}%
  \fontsize{9}{9.661}\selectfont
  \setlength{\itemsep}{0pt}%
  \setlength{\parsep}{0pt}%
  \setlength{\parskip}{0pt}%
}
\makeatother

\usepackage{cite}
\usepackage{amsmath,amssymb,amsfonts}

\usepackage{algorithmic}
\usepackage{graphicx}
\usepackage{textcomp}
\usepackage{xcolor}
\usepackage{soul}
\sethlcolor{yellow}
\usepackage{subfigure}
\usepackage{hyperref}
\usepackage[all]{hypcap}
\usepackage{verbatim}
\usepackage{media9}
\usepackage{balance}
\usepackage{makecell}
\usepackage{tabularx}
\usepackage{multirow}
\usepackage[T1]{fontenc}
\usepackage[utf8]{inputenc}
\usepackage{booktabs}
\usepackage{array}
\usepackage{url}
\usepackage{tikz}
\usetikzlibrary{arrows.meta,positioning,fit,shapes.geometric,backgrounds}

\def\L{{\cal L}}

\title{Low-Power, Neuromorphic, Acoustic Anomaly Detection for Persistent Machine Monitoring}
\name{%
\parbox{\textwidth}{\centering
\href{https://orcid.org/0000-0002-2327-7788}{Steven C. Nesbit}$^*$ \qquad 
\href{https://orcid.org/0000-0001-9404-0924}{Victor M. Vergara}$^{\dagger}$ \qquad
\href{mailto:michael.felix@cosmiac.org}{Michael A. Felix}$^{\ddagger}$ \qquad
\href{mailto:evan.kain.1@spaceforce.mil}{Evan T. Kain}$^{\mathsection}$\\
\href{https://orcid.org/0000-0003-0713-512X}{Luis R. Garc\'ia Carrillo}$^{\mathsection}$ \qquad
\href{https://orcid.org/0000-0002-4672-9484}{Gerd J. Kunde}$^{\mathparagraph}$ \qquad
\href{https://orcid.org/0000-0001-8036-6624}{Andrew T. Sornborger}$^*$}}

\address{%
\fontsize{9}{12.22}\selectfont
\parbox{0.95\textwidth}{\centering
$^*$Information Sciences (CAI-3), Los Alamos National Laboratory,
Los Alamos, NM, USA\\
$^{\dagger}$AeroVironment, Inc, Albuquerque, NM, USA\\
$^{\ddagger}$University of New Mexico COSMIAC Research Center, Albuquerque, NM, USA\\
$^{\mathsection}$Air Force Research Laboratory, Kirtland AFB, NM, USA\\
$^{\mathparagraph}$Nuclear \& Particle Physics \& Applications (P-3),
Los Alamos National Laboratory, Los Alamos, NM, USA\\
Email: $^*$\href{mailto:nesbitsc@lanl.gov}{nesbitsc@lanl.gov}}}

\begin{document}
\ninept
\maketitle
\begin{abstract}
Persistent acoustic monitoring can detect machine faults without physical contact, but always-on inference is constrained by power, latency, and deployment complexity. We demonstrate autoencoder-based acoustic anomaly detection on an Intel Loihi~2 neuromorphic processor under clean and noisy conditions. Log-mel features are computed off chip; normalization, autoencoder inference, L1 reconstruction scoring, and thresholding run on chip. In a clean, microphone-position-invariant ToyADMOS ToyCar benchmark, the on-chip model achieves 0.9959 AUC and 0.9785 standardized pAUC at maximum false-positive rate 0.1. In the DCASE 2026 Task~2 ToyCar noisy benchmark, the model achieves source AUC 0.7990, target AUC 0.6466, and pAUC 0.6426, exceeding reported baseline metrics. Power profiling on a 16-chip Loihi~2 VPX system shows real-time throughput with 0.0406--0.0426 mJ dynamic energy per sample, two orders of magnitude lower than both a CPU and GPU. These results support neuromorphic acoustic anomaly detection as a practical candidate for low-power, persistent machine monitoring.
\end{abstract}
\section{Introduction}
\label{sec:intro}

Persistent machine monitoring is an always-on sensing problem. A detector may spend most of its lifetime observing nominal operation, yet it must remain active so that rare faults are identified quickly. Acoustic monitoring is attractive because microphones can observe rotating and vibrating equipment without contact, while changes in motors, bearings, gears, and related components often produce measurable changes in spectral content.

Anomaly detection has been studied across statistics, data mining, machine learning, and signal processing, with applications including fraud, cybersecurity, medicine, industrial inspection, and spacecraft health monitoring \cite{chandola2009anomaly,ruff2021unifying,pang2022deep}. The literature includes supervised, semi-supervised, and one-class formulations. Supervised fault classifiers learn from labeled normal and anomalous examples, whereas one-class methods learn nominal behavior and score deviations without requiring anomalous training samples. Classical approaches include one-class support vector machines, support vector data description, local outlier factor, and isolation forests \cite{scholkopf1999support,tax2004support,breunig2000lof,liu2008isolation}. Deep approaches include reconstruction-based autoencoders, deep one-class objectives, latent density models, recurrent probabilistic models, adversarial autoencoders, and attention-based time-series detectors \cite{sakurada2014autoencoder,ruff2018deep,zong2018deep,su2019omni,audibert2020usad,xu2022anomalytransformer}. Supervised anomaly detectors can be effective when faults are known, but they solve a different problem and do not remove the need to detect previously unseen failures \cite{pang2021supervised}.

Machine-condition monitoring is particularly suited to one-class detection because failures are rare, heterogeneous, and expensive to collect. ToyADMOS, MIMII, and the DCASE anomalous sound detection tasks established public benchmarks in which models are trained primarily or exclusively on normal machine sounds \cite{ToyADMOS,purohit2019mimii,nishida2026dcase}. Prior acoustic work has investigated reconstruction, density-estimation, interpolation, and compact edge-deployable autoencoders \cite{abbasi2021outliernets,purohit2020dagmmacoustic,suefusa2020interpolation}. Related time-series work includes NASA/JPL spacecraft-telemetry detection using LSTM forecasting and dynamic thresholds \cite{hundman2018spacecraft}. These methods produce continuous deviation scores rather than requiring a separate output class for every possible fault.

Neuromorphic processors provide an alternative architecture for persistent inference through distributed state, sparse communication, and low-precision event-driven computation \cite{roy2019towards}. Intel's Loihi neuromorphic processor introduced programmable many-core neuromorphic processing with on-chip learning support \cite{davies2018loihi}. Loihi~2 adds flexible neuron dynamics, greater resource density, and integer-valued spike payloads for spiking, signal-processing, and fixed-point arithmetic workloads \cite{intel2021loihi2,orchard2021loihi2,shrestha2024efficient,davies2021advancing}. Neuromorphic anomaly detection has been studied using memristor and spiking autoencoders, converted spiking networks, evolving and balanced spiking networks, hierarchical temporal memory, homeostatic plasticity, and radiation-monitoring systems \cite{alam2019memristor,stratton2020spiking,jaoudi2020conversion,maciag2021unsupervised,dennler2021online,malawade2021neuroscience,nawaiseh2024homeostatic,ghawaly2022radiation}. Loihi has also been used for unsupervised automotive Controller Area Network bus anomaly detection, while TinySNN demonstrated unsupervised anomaly detection for industrial sensor streams \cite{islam2024unsupervised,mehrabi2024tinysnn}. Some acoustic SNN studies instead use labeled target events and are more accurately described as supervised event classification \cite{kshirasagar2024impact}. Despite this growing literature, little prior work has evaluated Loihi~2 for one-class acoustic machine monitoring with measured latency, power, energy, memory, and activity against conventional processors.

This study performs one-class anomaly detection in the strict sense: both autoencoders are trained exclusively on normal machine sounds, and anomalous recordings are used only for evaluation. Log-mel features are streamed to a Loihi~2 dense autoencoder, while normalization, reconstruction, L1 scoring, and thresholding are performed on chip. Our contributions are: (i) neuromorphic on-chip autoencoder inference and anomaly decisions; (ii) evaluation on clean ToyADMOS and noisy DCASE 2026 ToyCar datasets; (iii) comparison with the DCASE baseline using AUC and standardized pAUC; and (iv) latency, power, energy, memory, and activity profiling against server-grade CPU and GPU implementations.

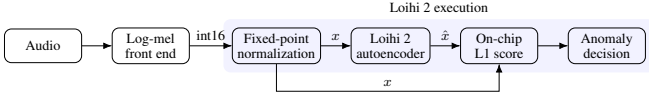
\begin{figure}[t]
\centering
\resizebox{\columnwidth}{!}{%
\begin{tikzpicture}[node distance=0.6cm,>=Latex,font=\footnotesize]
\tikzstyle{block}=[draw,rounded corners,align=center,minimum height=0.68cm,minimum width=1.55cm]

\node[block] (audio) {Audio};
\node[block,right=of audio] (mel) {Log-mel\\front end};
\node[block,right=0.85cm of mel] (norm) {Fixed-point\\normalization};
\node[block,right=of norm] (ae) {Loihi~2\\autoencoder};
\node[block,right=of ae] (score) {On-chip\\L1 score};
\node[block,right=of score] (decision) {Anomaly\\decision};

\begin{scope}[on background layer]
\node[fill=blue!5,rounded corners,fit=(norm)(ae)(score)(decision),inner sep=0.17cm,label={[yshift=0.03cm]above:{Loihi~2 execution}}] {};
\end{scope}

\draw[->] (audio) -- (mel);
\draw[->] (mel) -- node[above,inner sep=3pt]{int16} (norm);
\draw[->] (norm) -- node[above,inner sep=3pt]{$x$} (ae);
\draw[->] (ae) -- node[above,inner sep=3pt]{$\hat{x}$} (score);
\draw[->] (norm.south) |- ++(0,-0.55) -| node[pos=0.25,above,inner sep=3pt]{$x$} (score.south);
\draw[->] (score) -- (decision);

\end{tikzpicture}%
}
\caption{Persistent-monitoring pipeline. Log-mel extraction is off chip; normalization, autoencoder inference, L1 scoring, and thresholding run on Loihi~2.}
\label{fig:pipeline}
\end{figure}

\section{Methods}

\subsection{Datasets}

For the clean setting, we used the ToyADMOS ToyCar dataset \cite{ToyADMOS}. This dataset consists of 48-kHz recordings of a toy car captured from four microphone positions. The ToyCar subset contains 21\,600 normal recordings collected under four operating cases, corresponding to all combinations of two motor types and two bearing types. It also contains 4\,236 anomalous recordings spanning 53 anomaly conditions. The training set consisted of 17\,280 normal-condition recordings, ensuring that synchronized recordings from different microphone positions were not present in both the training and test sets. The test set contained 4\,320 normal and 4\,236 anomalous recordings.

For the noise-present dataset, we used the ToyCar subset of the DCASE 2026 Task 2 dataset, specifically the corrected ToyCar development release \texttt{dev\_ToyCar\_r2.zip} \cite{nishida2026dcase}. This dataset consists of 16-kHz, 10- to 12-second recordings of a toy car in a noisy environment. Each recording contains two synchronized channels: one recorded near the car and one recorded farther away. The far-channel recordings are provided as a potential noise reference; however, we used only the near-channel recordings to reflect a deployment scenario in which a separate noise-reference microphone may not be available. The training set contains 990 normal recordings from the source domain and 10 normal recordings from the target domain. The target domain differs from the source domain in factors such as operating speed, machine load, viscosity, heating temperature, environmental noise type, and signal-to-noise ratio. The test set contains 100 source-domain recordings, consisting of 50 normal and 50 anomalous recordings, and 100 target-domain recordings, also consisting of 50 normal and 50 anomalous recordings.

\subsection{Feature representation}

The input to the neuromorphic model is a log-mel representation of each audio segment: short-time spectral power mapped to mel-spaced frequency bands and logarithmically compressed. For the clean benchmark, recordings from the four ToyADMOS ToyCar microphone positions were pooled so that a single model was trained and evaluated across microphone-position variation. For the noise-present benchmark, feature extraction was performed only on the near-channel DCASE recordings.

For the clean 48-kHz ToyADMOS recordings, each audio file was converted to a log-mel spectrogram using 400 mel-frequency bins. Spectrograms were computed with a 2048-point FFT, a Hann analysis window, and a hop size of 512 samples. The mel spectrogram magnitudes were converted to decibel scale using the maximum power in each recording as the reference. To obtain a fixed-size input, a 93-frame segment was extracted from the middle portion of each spectrogram, corresponding to approximately one second of audio. The 93 frames were then averaged over time to produce a 400-dimensional log-mel vector for each recording.

For the noise-present 16-kHz recordings, the central one-second segment of each recording was converted to a compact log-mel feature vector. The short-time Fourier transform was computed using a 1024-point FFT, a Hann analysis window, and a hop size of 512 samples. The resulting power spectra were projected onto a 416-bin mel filterbank and transformed using the natural logarithm. The resulting log-mel spectrogram was then averaged over time to produce a fixed-length 416-dimensional feature vector.

\subsection{Autoencoder and on-chip score}

Both conditions use the same methodology: train an autoencoder on normal log-mel features, compute Z-score statistics from the normal training set, convert the model to fixed-point integer arithmetic, and use reconstruction error as the anomaly score. In fixed-point arithmetic, real values are stored as integers with fixed binary scale factors, enabling rescaling through multiplication and bit shifts rather than floating-point operations. Our autoencoders have layer widths

\begin{equation}
x \rightarrow 256 \rightarrow 128 \rightarrow 64 \rightarrow b \rightarrow 64 \rightarrow 128 \rightarrow 256 \rightarrow \hat{x}
\end{equation}

with $x$ representing the log-mel input, $b$ representing the dimensionally reduced bottleneck layer, and $\hat{x}$ representing the reconstruction. The bottleneck layer consisted of 32 neurons for the clean-data autoencoder and 12 neurons for the noisy-data autoencoder. The clean-data autoencoder used an input size of 400, and the noisy-data autoencoder used an input size of 416. The autoencoder used leaky-ReLU activations after hidden layers and a linear output layer. The trained models were exported to Loihi~2-compatible integer operations. Weights were represented as signed 8-bit values, while dense-layer accumulators, biases, and activations were constrained to signed 24-bit ranges. For each layer, the minimum arithmetic right shift required to keep all integer variables within the signed 24-bit range was used.

As illustrated in Fig. \ref{fig:pipeline}, the autoencoder output and normalized input are compared on chip. Because quantized input and reconstruction may use different scale factors, the L1 score is computed after independent rescaling:
\begin{equation}
 s = \sum_i \left| \left(\frac{m_x x_i}{2^{q_x}}\right) - \left(\frac{m_r \hat{x}_i}{2^{q_r}}\right) \right| 2^{-q_e}
 \label{eq:l1}
\end{equation}
where $x_i$ is the normalized log-mel input, $\hat{x}_i$ is the reconstruction, $m_x,m_r$ are integer multipliers, and $q_x,q_r,q_e$ are right bit shifts. The division notation in Equation (\ref{eq:l1}) is implemented by integer multiplication and arithmetic right shift. The final score can be emitted directly by Loihi~2 or compared to a threshold on chip for a binary anomaly decision.

Both floating-point models were trained for 280 epochs using Adam with learning rate \(10^{-3}\), batch size 16, and random seed 1031. The objective was mean-squared reconstruction error plus \(10^{-4}\) times the mean absolute bottleneck activation. No validation split, early stopping, dropout, or batch normalization was used. Because labeled DCASE development data informed model configuration, our DCASE results represent development-set rather than unseen challenge-evaluation performance.

Loihi~2 execution time, core allocation, memory utilization, and activity counts were obtained using on-board probes, while VPX input power was measured using a Keysight N6705C DC Power Analyzer. CPU execution time and package power were measured using wall-clock timing and Linux RAPL, while GPU execution time and board power were measured using CUDA events and 200-ms \texttt{nvidia-smi} sampling. Each workload was executed for 1\,000\,000 measured inferences following 1\,000\,000 warm-up inferences. Log-mel feature extraction was excluded from all measurements. Loihi~2 was evaluated with Ethernet communication and with preloaded inputs. The CPU and GPU implementations executed the same quantized autoencoder using int32 arithmetic at batch size 1.

\subsection{Evaluation metrics}

All models were evaluated using continuous anomaly scores, where larger scores indicate greater deviation from the normal training distribution. For each test recording, the anomaly score was computed from the central one-second segment.

Threshold-independent performance was measured using the area under the receiver operating characteristic curve (AUC). AUC evaluates how well the continuous anomaly scores rank anomalous recordings above normal recordings across all possible thresholds. It is appropriate here because deployment thresholds depend on the acceptable false-alarm cost and labeled anomalies are unavailable during training. We also report standardized partial AUC (pAUC) at a maximum false-positive rate of 0.1 following the DCASE convention \cite{nishida2026dcase}. Whereas AUC measures overall score separability, pAUC emphasizes the low-false-positive-rate region most relevant for persistent monitoring. For DCASE ToyCar, source AUC compares source-domain normal recordings against all anomalies, while target AUC compares target-domain normal recordings against all anomalies. ToyADMOS has no source and target domains, so pooled AUC and pAUC are reported across microphone positions.

\section{Results}

\subsection{Detection accuracy}

Table~\ref{tab:accuracy} summarizes the on-chip detection results. The clean benchmark demonstrates that the quantized Loihi~2 autoencoder provides strong separation between normal and anomalous log-mel features: the deployed L1 score produces AUC $0.9959$ and standardized pAUC $0.9785$. On the DCASE 2026 ToyCar benchmark, the Loihi~2 model exceeds the baseline on all three reported metrics. The strongest gain is on target AUC, where the on-chip model improves from $0.5317$ to $0.6466$. This result is notable because only the close microphone is used, as was done for the DCASE baseline autoencoder model; the far microphone is not used to characterize the noise field.

\begin{table}[t]
\centering
\caption{Anomaly-detection performance. pAUC is standardized pAUC at maximum FPR 0.1.}
\label{tab:accuracy}
\resizebox{\columnwidth}{!}{%
\begin{tabular}{lccc}
\toprule
Condition & Source/Pooled AUC & Target AUC & pAUC \\
\midrule
ToyADMOS ToyCar, Loihi~2 & 0.9959 & -- & 0.9785 \\
DCASE 2026 ToyCar, Loihi~2 & 0.7990 & 0.6466 & 0.6426 \\
DCASE 2026 ToyCar, baseline & 0.7728 & 0.5317 & 0.5825 \\
\bottomrule
\end{tabular}%
}
\end{table}

\subsection{Latency, power, and energy}

Table~\ref{tab:power} reports latency, mean power, and energy measured for autoencoder inference excluding feature extraction on a 16-chip Loihi~2 VPX system, a Xeon E5-2660 v3 CPU, and a Tesla V100S GPU. On the Loihi~2 VPX system, we report two measurements: a communication-inclusive measurement that includes Ethernet I/O and a second measurement with inputs preloaded to isolate the autoencoder execution time. With Ethernet I/O included, Loihi~2 processes one sample in $281.0~\mu$s, which is well below the audio-window rate required for real-time monitoring. When Ethernet I/O is excluded, the Loihi~2 autoencoder executes in $48.3~\mu$s per sample, faster than both the CPU and GPU measurements. Because GPU host-device transfer was excluded, the comparison does not disadvantage the GPU through communication overhead.

\begin{figure}[t]
\centering
\includegraphics[width=\linewidth]{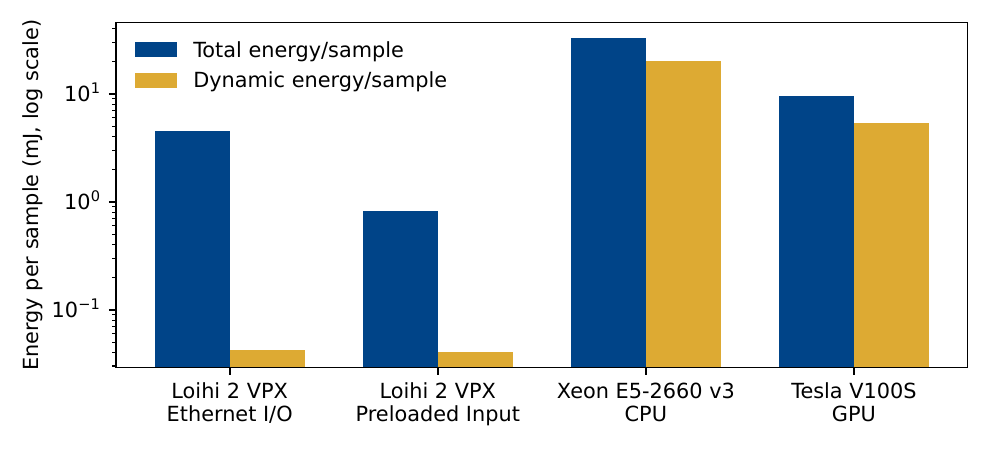}
\caption{Total and dynamic energy per sample; dynamic energy excludes static platform power.}
\label{fig:energy}
\end{figure}

The dynamic energy comparison in Fig.~\ref{fig:energy} indicates that Loihi~2 requires \(0.0426\) mJ of dynamic energy per sample with Ethernet I/O and \(0.0406\) mJ with preloaded inputs, compared with \(20.16\) mJ on the CPU and \(5.35\) mJ on the GPU. The communication-inclusive Loihi~2 measurement therefore uses approximately \(474\times\) less dynamic energy than the CPU and \(126\times\) less than the GPU; with preloaded inputs, the reductions are approximately \(496\times\) and \(132\times\), respectively. Thus, the neuromorphic implementation retains a substantial energy advantage even when Ethernet I/O is included.

The mapped autoencoder occupied 74 neuromorphic cores. The on-board probes measuring per-core memory usage reported single-chip utilization of 20.36\%, with the corresponding memory utilization across the 16-chip VPX system being 1.27\%. Per inference, the activity probes recorded approximately \(3.0\times10^5\) synaptic operations, \(1.0\times10^4\) neuron updates, \(2.0\times10^4\) input spike events, and \(2.0\times10^4\) output spike events. The 74-core allocation and 20.36\% aggregate single-chip memory utilization are both within the capacity of one 128-core Loihi~2 chip, although a direct single-chip deployment would still be needed to verify routing and maximum per-core resource utilization. The mapped model is therefore compatible in core count with a single-chip Loihi~2 system such as Intel's Oheo Gulch platform \cite{intel2021loihi2}. Although Oheo Gulch was not available for direct measurement, prior work reports static power of \(1.56~\mathrm{W}\) for its x86 subsystem and \(0.22~\mathrm{W}\) for its neuromorphic cores \cite{NesbitStevenChristian2025Aobf}. Scaling these values to the VPX embedded processor subsystem and the 16 Loihi~2 chips of the VPX board gives
\begin{equation}
    8(1.56~\mathrm{W}) + 16(0.22~\mathrm{W}) = 16.00~\mathrm{W}
\end{equation}

which closely matches the measured VPX static power of \(16.05~\mathrm{W}\). Using \(1.78~\mathrm{W}\) as the estimated single-chip static power and retaining the measured VPX dynamic energy and latency gives projected total powers of \(1.93~\mathrm{W}\) with Ethernet I/O and \(2.62~\mathrm{W}\) with preloaded inputs. The corresponding projected total energies are \(0.54~\mathrm{mJ}\) and \(0.13~\mathrm{mJ}\) per sample. These values represent first-order projections that assume comparable dynamic and communication behavior on the smaller platform; direct single-chip measurements are required for validation.

\begin{table}[t]
\centering
\caption{Inference latency and energy profiling.}
\label{tab:power}
\resizebox{\columnwidth}{!}{%
\begin{tabular}{lcccc}
\toprule
Hardware & \makecell[c]{Latency} & \makecell[c]{Total power} & \makecell[c]{Total energy} & \makecell[c]{Dynamic energy} \\
 & \makecell[c]{($\mu$s/sample)} & \makecell[c]{(W)} & \makecell[c]{(mJ/sample)} & \makecell[c]{(mJ/sample)} \\
\midrule
Loihi~2 VPX, Ethernet I/O & 281.0 & 16.20 & 4.554 & 0.0426 \\
Loihi~2 VPX, Preloaded Input & 48.3 & 16.89 & 0.816 & 0.0406 \\
Xeon E5-2660 v3 CPU & 420.3 & 77.96 & 32.769 & 20.156 \\
Tesla V100S GPU & 163.5 & 58.35 & 9.540 & 5.350 \\
\bottomrule
\end{tabular}%
}
\end{table}

\section{Discussion}

Our results support Loihi~2 as a practical substrate for persistent acoustic anomaly detection. The clean benchmark demonstrates strong separation under microphone position variation, while the DCASE result remains effective in noisy conditions using only the near microphone and no separate noise-reference channel. The AUC and pAUC values show that the neuromorphic model provides useful anomaly-score ranking across both monitoring conditions.

Although static platform power dominates the 16-chip VPX measurement, dynamic energy per sample is low and the model occupies only 74 of approximately 2048 available neuromorphic cores. This leaves substantial capacity for additional models, sensor fusion, or future log-mel preprocessing.

The principal limitation is that log-mel extraction remains off chip. Although Ethernet-I/O timing is compatible with real-time monitoring, on-chip preprocessing could further reduce host dependence and total system energy. Future work should also evaluate additional DCASE machine classes and field recordings, directly profile a single-chip Loihi~2 platform, and investigate online adaptation under changing machine and environmental conditions.

\section{Conclusion}
This paper demonstrates a Loihi~2 neuromorphic implementation of log-mel autoencoder anomaly detection for clean and noisy acoustic monitoring. The on-chip system achieves 0.9959 AUC and 0.9785 pAUC in the clean benchmark and exceeds the DCASE 2026 ToyCar baseline with source AUC 0.7990, target AUC 0.6466, and pAUC 0.6426. Power profiling shows real-time throughput and very low dynamic energy per sample compared with CPU and GPU inference. These results indicate that neuromorphic autoencoders are a promising path toward low-power, persistent acoustic machine monitoring, with on-chip log-mel preprocessing and direct single-chip Loihi~2 measurements as important next steps.

\section{Acknowledgments}

This work was supported by the US Department of Energy National Nuclear Security Administration's Office of Defense Nuclear Nonproliferation Research \& Development (DNN R\&D) at Los Alamos National Laboratory under contract 89233218CNA000001. The authors' employment affiliations are listed in the author block. We thank AFRL/RJSV's SPACER lab for providing the facilities and support necessary for this research. The views expressed are those of the author and do not necessarily reflect the official policy or position of the Department of the Air Force, the Department of Defense, or the U.S. government. Distribution Statement A: Approved for Public Release. Distribution is Unlimited. Public Affairs Release Approval \#AFRL-2026-3570; Los Alamos National Laboratory approval \#LA-UR-26-26527.

\section{Compliance with Ethical Standards}

This study used publicly available machine-sound datasets and therefore ethical approval was not required.

%\vfill\pagebreak

%Please follow the IEEE Citation Guidelines, \url{https://ieee-%dataport.org/sites/default/files/analysis/27/IEEE\%20Citation\%20Guidelines.pdf} for formatting of references.


\bibliographystyle{IEEEbib}
\bibliography{strings,refs}

\end{document}